# Overlapping clustering based on kernel similarity metric

Chiheb Eddine Ben N'Cir[1], Nadia Essoussi[1], Patrice Bertrand[2]

[1] *Larodec, ISG University of Tunis*

[2] *Ceremade, Université Paris-Dauphine*

*ABSTRACT: Producing overlapping schemes is a major issue in clustering. Recent proposed overlapping methods relies on the search of an optimal covering and are based on different metrics, such as Euclidean distance and I-Divergence, used to measure closeness between observations. In this paper, we propose the use of another measure for overlapping clustering based on a kernel similarity metric .We also estimate the number of overlapped clusters using the Gram matrix. Experiments on both Iris and EachMovie datasets show the correctness of the estimation of number of clusters and show that measure based on kernel similarity metric improves the precision, recall and f-measure in overlapping clustering.*

KEYWORDS: Overlapping clustering, kernel method, gram matrix, similarity measure.

## 1. Introduction

Overlapping clustering is a major issue in clustering. It aims to divide the data into similar groups in which an object can belong to one or more clusters without any membership coefficent. Based on the assumption that an object really belongs to many clusters, overlapping clustering is different from crisp and fuzzy clustering. The problem of finding overlapping schemes has been gaining importance in a wide variety of application domains because many real problems are naturally overlapped. For example, in social network analysis, community extraction algorithms should be able to detect overlapping clusters because an actor can belong to multiple communities. Also, in computational biology, overlapping clustering is a necessary requirement in the context of microarray analysis and protein function prediction because protein can potentially have multiple functions.

Several overlapping clustering methods based on stratified and partitioning approaches are proposed in the literature. Stratified methods are consisted in pyramids [DID 84], which are structures less restrictive than the trees, and k-weak hierarchies [BER 03], which are a generalization of the pyramids. These methods have a major issue that corresponds to the limitation of the overall patterns of possible covering.

Overlapping methods based on partitioning approach extended primarily the methods of strict or fuzzy classification to produce overlapped clusters. Several clustering methods have been used such as soft k-means and Threshold Meta-clustering Algorithm [DEO 06]. The main issue in these methods is the prior threshold which is difficult to learn.

Early researches in partitioning methods solve the thresholding problem. Cleuziou [CLE 04] proposed PoBOC (Pole Based Overlapping Clustering) where individuals are assigned to clusters by studying the distribution of their proximity to all clusters. A variant of this algorithm called DDOC (Distributional Divisive Overlapping Clustering) [CLE 04] treats text data which are characterized by their large size.

The criteria optimized successively in these partitioning methods looks for an optimal partition without introducing the overlap between data in optimization step. More recent models on overlapping clustering solve this problem and look for optimal covering. Banerjee [BAN 05] proposed MOC (model based overlapping clustering) which is considered as the first algorithm looking for optimal covering. This algorithm is inspired from biology and is based on the PRM (Probabilistic Relational Model). Cleuziou [CLE 07] proposed OKM (Overlapping K-Means) which is considered as a generalization of k-means to detect overlap between data.

Cleuziou [CLE 09] proposed also the OKMED method (Overlapping K-Medoid) which generalizes the method PAM (Partionning Around Medoid). Instead of computing cluster's center, OKMED represents each cluster with an object among the objects to be clustered.

Other approaches were recently used to produce overlapping algorithms such as graph theory based approach and neural network. Methods based on graph theory begin by building a similarity graph. Then, from this graph, it seeks for the maximum number of cliques. Didimo [DID 07] proposed the method OCP (overlapping clustering planarity) and Fellows [FEL09] proposed the method Graph-Based Data Clustering with Overlaps. A new method proposed recently by Cleuziou [CLE 10] generalizes the method SOM (Self Organizing Map) and is based on neural approach. It is based on models of self-organization to produce Overlapping patterns.

In this paper, we propose an approach to estimate the number of covering which can be very useful for parametric overlapping method such as OKM, OKMED and MOC. We also propose another similarity measure between objects based on kernel similarity metric which can improve overlapping clustering results.

This paper is organized as follows: Section 2 presents the used method to estimate the number of clusters and the kernel metric used to improve overlapping clustering, Section 3 describes experiments and results on IRIS data set and EachMovie subset. Finally, Section 4 presents conclusion and future works.

## 2. Methodology

The overall objective of this study is to improve overlapping clustering quality. For this purpose, two advantages of kernel methods are carried to improve results. Firstly, we use the Gram matrix to estimate the number of covering. Secondly, we use each value in the Gram matrix as a similarity measure between objects. A comparison between kernel based similarity measure, Euclidean distance and I-Divergence is done in terms of recall, precision and f-measure.

### 2.1 Estimating the number of covering

In the case of overlapping schemes, the overlap between clusters is an important characteristic that affects the determination of the appropriate number of clusters. However, it remains difficult to give a prior preference between an organization with some clusters with strong overlap or an organization with many clusters which are smaller overlapped. In general, the number of covering is the same number of partitions.

For this purpose, Kernels have proven to be extremely powerful in many areas of machine learning [SCH 02] [BEN 00]. The activities regarding the study of Kernel Methods for Clustering are still in progress. The goal of these activities is to yield kernel algorithms, whose convergence is proved. Kernel is defined as a function K such that $K(x_i, x_j) = <\phi(x_i), \phi(x_j)>$ for all $x_i, x_j \in X$ where $\phi$ is a mapping from X to feature space F. A kernel matrix (Gram matrix) is a square matrix $K \in R^{n*n}$ such that $K(i,j)$ is equal to $K(x_i, x_j)$ for all $x_1,...x_n \in X$ and some kernel function K. There exist many kernel functions in the literature and mostly used are:

RBF kernel: $\exp\left\{\dfrac{-\|x_i - y_i\|^2}{\sigma^2}\right\}$      Polynomial kernel: $(<x, y> + 1)^d$

The kernel matrix (Gram matrix) can be used to determine the number of clusters in data set. As each element of the kernel matrix defines a dot-product distance in the feature space, the matrix will have a block diagonal structure when there are definite clusters within the data sets. This diagonal structure block can be used to determine the number of clusters. This method was first used in the c-means clustering, and it still interesting in overlapping clustering. Thus, through counting the number of significant Eigen values of kernel matrix, we can obtain the number of clusters.

## 2.2 Kernel based similarity measure

Mercer Kernel functions map data from input space to high, possibly infinite, dimensional feature space. For a finite sample of data X, the kernel function yields a symmetric N x N positive definite matrix K, where the K(i, j) entry corresponds to the dot product between $\phi(x_i)$ and $\phi(x_j)$ as measured by the kernel function.

In feature space, the distance measure between any two patterns is given by:

$$\|\phi(x_i) - \phi(x_j)\|^2 = <\phi(x_i),\phi(x_i)> + <\phi(x_j),\phi(x_j)> - 2<\phi(x_i),\phi(x_j)> \quad [1]$$

$$\|\phi(x_i) - \phi(x_j)\|^2 = K(i,i) + k(j,j) - 2k(i,j)$$

If the kernel used is an RBF kernel, the function [1] becomes:

$$\|\phi(x_i) - \phi(x_j)\|^2 = 2 - 2\exp\left\{-\frac{\|x_i - x_j\|^2}{\sigma^2}\right\} \quad [2]$$

In addition, the objective function should be adapted to look for optimum in feature space. For example, in OKM the objective function to minimize is as follows:

$$J(\{\pi_c\}_{c=1}^k) = \sum_{x_i \in X} \|x_i - image(x_i)\|^2 \quad [3]$$

Where $X = \{x_i\}_{i=1}^n$ is the data vector set with $x_i \in R^P$ and $\{\pi_c\}_{c=1}^k$ is the set of k covering.

The function [3] minimizes the distance between each object and its corresponding image. Then, using kernel similarity measure, this objective function is modified as follows:

$$J(\{\pi_c\}_{c=1}^k) = \sum_{x_i \in X} \|\phi(x_i) - \phi(image(x_i))\|^2 \quad [4]$$

$$= \sum_{x_i \in X} [\phi(x_i) * \phi(x_i)] + [\phi(image((x_i))) * \phi(image((x_i)))] - 2[\phi(x_i) * \phi(image((x_i)))]$$

Then, each dot product in feature space is replaced by the function of kernel trick:

$$J(\{\pi_c\}_{c=1}^k) = \sum_{x_i \in X} [K((x_i),(x_i))] + [K(image(x_i), image(x_i))] - 2[K((x_i), image(x_i))] \quad [5]$$

If the kernel used is an RBF kernel, the function [5] becomes:

$$J(\{\pi_c\}_{c=1}^{k}) = \sum_{x_i \in X} \left(2 - 2\exp\left\{\frac{-\|x_i - image(x_i)\|^2}{\sigma^2}\right\}\right) \quad [6]$$

## 3. Experiments

Experiments are performed on both Iris and EachMovie datasets by using OKM with kernel similarity metric and OKM with different distance measures such as Euclidean distance and I-Divergence. The estimation of the number of clusters is obtained by building the gram matrix from these datasets and by extracting respectively the most significant Eigenvalues.

### 3.1 Iris data set

The Iris dataset is traditionally used as a base's test for evaluation. It is composed of 150 data in $R^4$ tagged according to three non-overlapping clusters (50 data per class). One of these clusters (setosa) is known to be clearly separated from the two others [WEB $10^2$].

Figure 1 shows the most significant Eigenvalues of the Gram matrix. The kernel used is RBF kernel with $\sigma$=150. There are at least 2 covering, another covering looks less important. If we choose to add this covering then we obtain less overlap between data. So, known that Iris doesn't contain overlaps between data, it's more suitable to add this third cluster. The optimal choice is then 3 clusters.

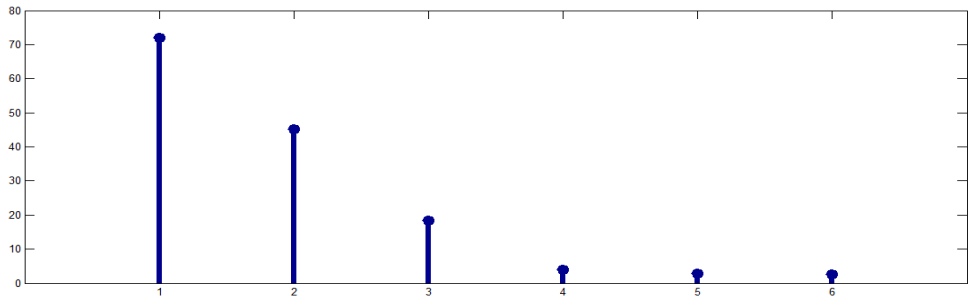

FIGURE 1 – most significant Eigenvalues in IRIS data set

Then using the Euclidean distance, the I-Divergence and the kernel similarity metric and by estimating the number of cluster to k = 3, we run OKM ten times (with similar initializations). The mean, the min and the max of results obtained are reported according to three validation measures: precision, recall, f-measure. For each pair of points sharing at least one cluster in the overlapping clustering results, these validation measures try to estimate whether the prediction of this pair was correct with respect to the underlying true categories in the data.

Precision is calculated as the fraction of pairs correctly put in the same cluster, recall is the fraction of actual pairs that were identified, and F-measure is the harmonic mean of precision and recall:

$$Precision = NCILP/NILP \quad [7]$$

$$Recall = NCILP/NTLP \quad [8]$$

$$F\text{-}measure = 2 * Precision * Recall / Precision + Recall \quad [9]$$

where NCILP, NILP and NTLP are respectively the number of correctly identified linked pairs, the number of identified linked pairs and the number of true linked pairs.

*TABLE 1* – *Comparison between different similarities measures used in overlapping k-means (OKM) on IRIS dataset*

| Distance measure | MIN | | | MAX | | | MEAN | | |
|---|---|---|---|---|---|---|---|---|---|
| | Precision | Recall | F-measure | Precision | Recall | F-measure | Precision | Recall | F-measure |
| Euclidean distance | 0.603 | 0.853 | 0.706 | 0.765 | 0.993 | 0.854 | 0.707 | 0.900 | 0.815 |
| I-Divergence | 0.514 | 0.733 | 0.617 | **0.848** | 1 | 0.906 | 0.759 | **0.993** | 0.834 |
| RBF kernel($\sigma$=150) | 0.585 | 0.906 | 0.711 | 0.782 | 0.993 | 0.901 | 0.771 | 0.960 | 0.830 |
| Polynomial kernel(d=0.25) | **0.771** | **0.936** | **0.865** | 0.838 | 1 | **0.911** | **0.808** | **0.993** | **0.892** |

Table 1 shows the usefulness of polynomial kernel used with OKM. Measure based on Polynomial kernel ameliorate results of overlapping clustering compared to measure based on Euclidean distance and I-divergence.

### 3.2 EachMovie dataset

The EachMovie dataset contains user ratings for every movie in the collection. Users give ratings on a scale of 1-5, with 1 indicating extreme dislike and 5 indicating strong approval. There are 74,424 users in this dataset, but the mean and median number of users voting on any movie are 1732 and 379 respectively [WEB 10[1]]. As a result, if EachMovie in this dataset is represented as a vector of ratings over all the users, the vector is high-dimensional but typically very sparse. For every movie in the EachMovie dataset, the corresponding genre information is extracted from the Internet Movie Database (IMDB) collection. If each genre is considered as a separate category or cluster, then this dataset has naturally overlapping clusters since many movies are annotated in IMDB as belonging to multiple genres. For example, Aliens movie belongs to 3 genre categories: action, horror and science fiction.

We extracted subset from the EachMovie dataset: 75 objects scattered on three overlapping clusters as follows: action=21objects; comedy=26 objects; crime=17 objects; action+crime=11 objects; And based on age, sex and rate of users we try to find a category of video.

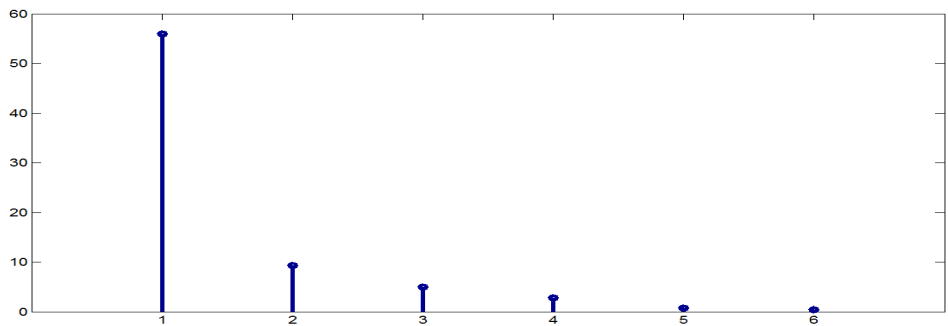

*FIGURE 2* – *most significant Eigenvalues in EachMovie dataset*

Figure 2 shows the most significant Eigenvalues of the Gram matrix. The kernel used is RBF kernel with σ=2. We have between 3 and 4 significant Eigenvalues. Known that EachMovie subset is an overlapping subset, the suitable choice is 3 clusters.

*TABLE 2 – Comparison between different similarities measures used with overlapping k-means on EachMovie subset*

| Distance measure | MIN | | | MAX | | | MEAN | | |
|---|---|---|---|---|---|---|---|---|---|
| | Precision | Recall | F-measure | Precision | Recall | F-measure | Precision | Recall | F-measure |
| Euclidean distance | 0.494 | 0.641 | 0.561 | 0.673 | **0.919** | 0.740 | 0.582 | 0.827 | 0.687 |
| I-Divergence | 0.541 | 0.552 | 0.540 | 0.532 | 0.814 | **0.814** | 0.49 | 0.687 | 0.603 |
| RBF kernel(σ=2) | 0.494 | 0.641 | 0.540 | 0.673 | **0.919** | 0.740 | 0.582 | 0.827 | 0.687 |
| Polynomial kernel(d=2) | **0.55** | **0.693** | **0.653** | **0.7** | 0.909 | 0.766 | **0.628** | **0.851** | **0.721** |

Then, using the Euclidean distance, I-Divergence and kernel similarity metric and by estimating the number of cluster to k = 3, we run OKM ten times (with similar initializations) and we report the mean, the min and the max of results obtained according to three validation measure: precision, recall, f-measure. Table 2 shows the usefulness of polynomial kernel used with OKM. RBF kernel gives the same result of Euclidean distance. These results confirm the first results obtained in Iris dataset. Measure based on Polynomial kernel improve results of overlapping clustering compared to measure based on Euclidean distance and I-Divergence.

## 4. Conclusion and Future works

In this paper, we have proposed a measure based on kernel method which is used in several hard and fuzzy clustering algorithms and it remains useful in overlapping clustering. We have also used a Gram matrix to estimate the number of covering. This estimation is useful for parametric methods which consider the number of cluster known before the run of the algorithm. Experiments on both Iris and EachMovie dataset showed an improvement of the results of overlapping clustering algorithm.

We plan to extend this work by considering another key advantage of kernel method to improve overlapping clustering quality. Kernel methods are used to find non-linear boundaries between clusters. This idea has been used in hard and soft clustering but is meaningful in overlapping clustering. In addition, when using kernel method, overlapping clustering can be applied to structured data, such as trees, strings, histogram and graphs.